\documentclass[runningheads]{llncs}

\usepackage{eccv}

\usepackage{eccvabbrv}
\usepackage{comment}
\usepackage{graphicx}
\usepackage{booktabs}
\usepackage{subcaption}
\usepackage{multirow}
\usepackage{colortbl}
\usepackage[accsupp]{axessibility}  

\usepackage[pagebackref,breaklinks,colorlinks,citecolor=eccvblue]{hyperref}

\usepackage{wrapfig}

\usepackage[ruled,vlined]{algorithm2e}
\usepackage{multirow}
\usepackage{bbm}
\usepackage[ruled,vlined]{algorithm2e}
\definecolor{commentcolor}{RGB}{110,154,155}  

\newcommand{\PyComment}[1]{\ttfamily\textcolor{commentcolor}{\# #1}}  
\newcommand{\PyCode}[1]{\ttfamily\textcolor{black}{#1}} 

\def\@fnsymbol#1{\ensuremath{\ifcase#1\or *\or \dagger\or \ddagger\or
   \mathsection\or \mathparagraph\or \|\or **\or \dagger\dagger
   \or \ddagger\ddagger \else\@ctrerr\fi}}
\newcommand{\ssymbol}[1]{^{\@fnsymbol{#1}}}
\usepackage{caption}
\usepackage[ruled,vlined]{algorithm2e}
\usepackage{multirow}
\begin{document}
\title{BD-KD: Balancing the Divergences for Online Knowledge Distillation} 

\titlerunning{Abbreviated paper title}

\author{Ibtihel Amara\inst{1} \and
Nazanin Sepahvand \inst{1} \and
Brett H. Meyer \inst{1} \and
Warren J. Gross \inst{1} \and
James J. Clark \inst{1}
}

\authorrunning{Amara et al.}

\institute{McGill University
}
\maketitle
\begin{abstract}
We address the challenge of producing trustworthy and accurate compact models for edge devices. While Knowledge Distillation (KD) has improved model compression in terms of achieving high accuracy performance, calibration of these compact models has been overlooked. We introduce BD-KD (Balanced Divergence Knowledge Distillation), a framework 
for logit-based online KD. BD-KD enhances both accuracy and model calibration simultaneously, eliminating the need for post-hoc recalibration techniques, which add computational overhead to the overall training pipeline and degrade performance. Our method encourages student-centered training by adjusting the conventional online distillation loss on both the student and teacher losses, employing sample-wise weighting of forward and reverse Kullback-Leibler divergence. This strategy balances student network confidence and boosts performance. Experiments across CIFAR10, CIFAR100, TinyImageNet, and ImageNet datasets, and various architectures demonstrate improved calibration and accuracy compared to recent online KD methods.
\end{abstract}

\section{Introduction}
\label{sec:intro}
A prime factor in the viability of numerous critical edge applications is the fusion of true reliability with exceptional accuracy. A compelling illustration of this can be witnessed in the realm of autonomous vehicles, a domain where the imperative of compactness is principal. In such cases, the deep neural networks responsible for intricate decision-making processes must be meticulously sized to seamlessly integrate with electronic chips having restricted computational power and limited memory capacity. The success of such applications demands maintaining a balance between compactness, precision, and calibrated predictive confidence for uncertainty.
Much research \cite{courbariaux2015binaryconnect,jacob2018quantization,han2015learning, guo2016dynamic, wu2016quantized,denton2014exploiting,kim2018paraphrasing} has focused on compressing large well-performing deep models with billions of parameters. Among the more successful compression techniques is knowledge distillation (KD), a teacher-student training paradigm. KD is known for its simplicity and versatility as it is architecture agnostic, with no constraints on the type of architecture of either the teacher or the student networks. However, KD techniques have been hampered by problems that arise when there is a large capacity gap between teacher networks and smaller student networks, as was first noted by \cite{mirzadeh2020improved}.

%
%
\setlength\intextsep{0pt}
\begin{figure}[t]
\begin{center}
\includegraphics[width=0.6\linewidth]{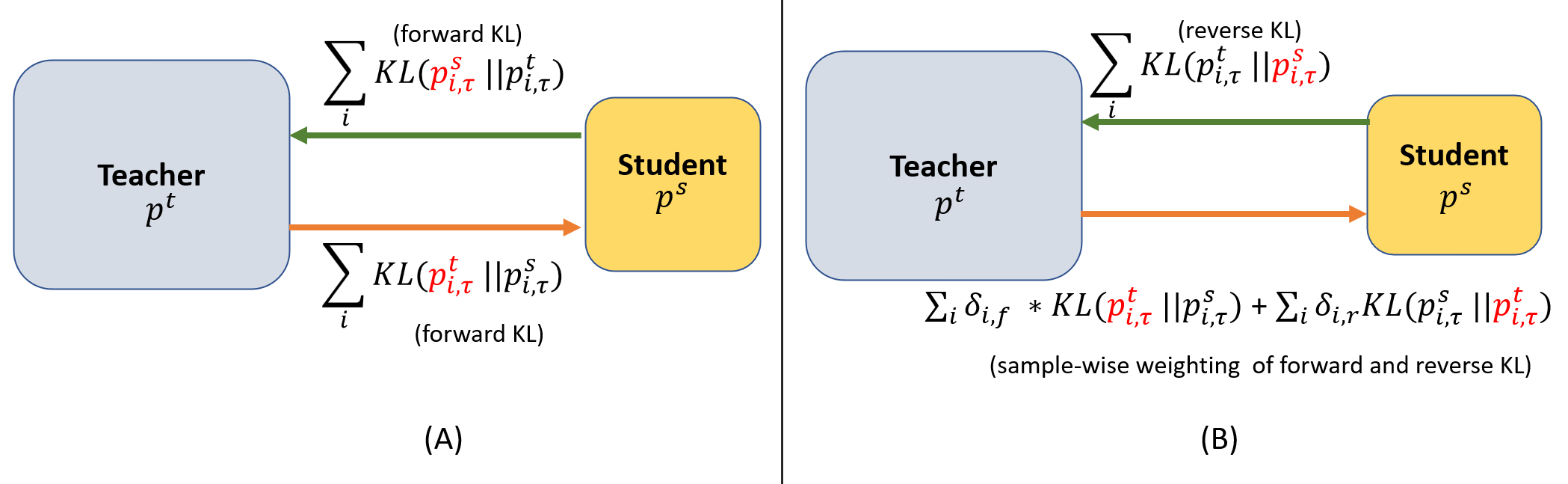}
\end{center}
\caption{
{\footnotesize \textbf{Distillation losses in the proposed framework.} (A) depicts the conventional online distillation loss \cite{dml}. (B) depicts the proposed student-centered distillation loss. The feedback signal from teacher to student takes in a sample-wise weighting of both forward and backward KL. We also exploit the reverse KL for teacher training. The parameters in red font detached (stop gradient flow) during training.
}}
\label{fig:framework}
\end{figure}
The predominant focus of knowledge distillation (KD) research has evolved around enhancing the student's accuracy, employing intricate design principles and multifaceted objective functions. 
However, insufficient attention has been directed toward the improvement of calibration in the resultant compact student model.
Calibration is associated with the dependability and trustworthiness of networks\cite{guo2017calibration}. A well-calibrated model is one where where the estimated confidence probabilities reflect the genuine probabilities of the associated outcomes. For example,  should the model provide an answer with a confidence estimate of 0.95, in a well-calibrated network this would signify that the predicted event materializes in reality 95\% of the time. This alignment between prediction and actuality lays the foundation for robust and enlightened decision-making, thus accentuating the model's reliability and efficacy.

In our study, we build upon recent empirical observations \cite{dml,switokd,li2022shake} that accentuate the efficacy of online distillation in augmenting student model accuracy. We focus on obtaining compact student models that are simultaneously well-performing and better-calibrated. 
We propose a straightforward learning framework for efficient online KD training called BD-KD: \textit{Balanced divergence Knowledge Distillation}. 
Our approach exploits the salient principle of collaborative training with a special focus on the student’s learning process, wherein both the teacher and student networks' training are orchestrated to engage in a reciprocal exchange of informative feedback signals via the integration of their Kullback-Leibler (KL) divergence loss. 

Our main contributions are: 
\vspace{-1.5mm}
\begin{itemize}
    \item We propose an online KD training strategy with a focus on student model training where the conventional forward KL loss in the student objective function is replaced with a sample-wise weighting of both forward and backward KL. We demonstrate that this weighting scheme significantly enhances the student’s performance via better output uncertainty estimation and calibration. 
    \item Our method outperforms state-of-the-art KD models, both online and offline, in terms of accuracy, for four different computer vision datasets and across multiple network architectures. 
    \item Our approach improves student network calibration, compared to both existing online and offline methods. To the best of our knowledge, we are the first to report the connection between online distillation and improved model calibration.
    \item Our approach can mitigate the capacity gap issue so that compact student networks profit from larger capacity teachers without performance loss.
\end{itemize}
The novelty of our paper is most evident in student-focused distillation, which sculpts and redefines both model compression and calibration paradigms. By leveraging a confidence-balancing mechanism during distillation training from an uncalibrated large teacher network, we achieve noteworthy performance. Furthermore, our framework readily integrates with diverse objective functions for various tasks and applications.

\section{Related works}
\underline{\textbf{Student-focused Online KD.}}
A prominent contribution in the domain of student-focused online KD was introduced by \cite{qian2022switchable}. Their method, called SwitOKD, attempts to enhance the student model accuracy by bridging the accuracy gap during training. This is ensured by strategically slowing down the training process of the teacher, enabling the student network to make substantial gains. This approach shifts the offline and online KD phases, infusing the training process with a student-centered focus. Similarly, \cite{li2022shake} propose a proxy teacher training paradigm, called SHAKE, where the weights are updated through the use of the teacher model (offline KD) and the student during online KD training. They believe that the proxy teacher is encouraged to be aware of the capacity of the student's training, which encourages student-centered training. 
While these methods set impressive benchmarks for online KD, they do come with some caveats. SwitOKD prolongs training due to the fact we are slowing down the training of the teacher network, and SHAKE adds complexity through the use of a proxy teacher and the proposed overall design.
In this paper, we propose a new online distillation method where, just like the aforementioned methods, we focus on having student-centered online KD training with the main goal of achieving high performance with improved calibration. Our new approach involves modification to the online KD objective functions, where both forward and reverse KL divergence losses are employed.
\\
\underline{\textbf{Symmetric KL Insights.}} Incorporating both forward and backward KL has demonstrated robust training and significant performance \cite{binici2022preventing,yin2020dreaming}. Certain data-free offline KD approaches have embraced the Jensen-Shannon divergence (JSD), harnessing forward and reverse KL divergences to synthesize non-redundant images more effectively. This approach also aids in quantifying teacher-student discrepancy in data-free scenarios \cite{binici2022preventing}. Lee et al. \cite{Lee2022SelfKnowledgeDV} employed the simple symmetric KL by combining both forward and reverse KL terms in their proposed offline self-distillation through dropout. 
Unlike JSD-based methods, our approach uniquely balances forward and backward KL divergences based on student-teacher uncertainty, adapting to mode-seeking and mean-seeking behavior. This crucial distinction, which sets our work apart and dynamically addresses student-teacher uncertainty estimation leading to significant calibration and performance improvements.\\
%
\underline{\textbf{Sample-wise re-weighting in KD.}} In KD, sample-wise re-weighting of the KD loss has proven highly efficient for effective distillation performance. Tang et al. \cite{tang2020understanding} showed that dynamically adjusting student model weights during training, based on a logit rescaling factor derived from the teacher's relative prediction confidence for each sample, significantly improves the student's overall performance. Similarly, Zhou et al. \cite{zhou2021rethinking} suggested a weighting method for soft labels, applying smaller weights if the student model outperforms the teacher model for a given sample. Their experiments demonstrated that sample-wise re-weighting enhances distillation performance. Su et al. \cite{su2023knowledge} proposed an entropy re-weighting strategy, employing the teacher's entropy to re-weight the student KD loss.

\noindent Our work differs substantially from the aforementioned sample-wise re-weighting techniques. Most of these techniques are used in the offline KD setting, utilizing a single term of KL divergence, usually the forward KL, and applying the sample-wise weighting factor solely on that term. In contrast, BD-KD harmonizes the contributions of both forward and reverse KL divergences within the KD loss by using the uncertainty gap (i.e entropy gap) between the large teacher and the compact student networks as a sample-wise weighting factor.
\\
\underline{\textbf{Calibration in KD.}} Model calibration has captured the attention of much research \cite{guo2017calibration, muller2019does, mukhoti2020calibrating, yun2020regularizing,zhong2021improving} as it addresses the uncertainty of deep networks given their empirical accuracy. There are many calibration techniques in the literature. Some focus on enhancing calibration through extensive data augmentation, such as mixup \cite{thulasidasan2019mixup}, while others focus on post-hoc techniques like temperature scaling \cite{guo2017calibration}, which add computation overhead to training and reduce performance. Recent research has shown that conventional distillation techniques encourage the student model to be better calibrated as compared to the same model trained from scratch \cite{yun2020regularizing, stanton2021does}. Current methods are shifting towards obtaining well-calibrated networks via KD. In our work, by adaptive weighting between the reverse and forward KL to balance the over-confidence and under-confidence of the compact student in training, making the distillation training more student-centric, we obtain better-calibrated compact student models without affecting their overall performance. 
%
\section{Methodology}
\label{sec:method}
Our framework is presented in Figure \ref{fig:framework}. In the original distillation method of \cite{hinton2015distilling}, soft labels obtained from a pre-trained teacher are used to train a student network. In this form of KD, the objective function used to guide the student's training consists of two terms: (1) classic cross-entropy loss, $L_{CE}$, that minimizes the distance between student's prediction and hard labels, and (2) the distillation loss, which consists of the KL divergence, $L_{Distill} = L_{KL}$, whose goal is to bring student's logits closer to that of the teacher.

The type of KL divergence used in most KD approaches is the \textbf{\emph{forward KL}}: \boldmath{$KL(p^t||p^s) = \sum_{x} p^t(x) \log\left(\frac{p^t(x)}{p^s(x)}\right)$}, where $p^t$ and $p^s$ are the softmax probabilities of the teacher and the student, respectively. 
One major shortcoming of forward KL divergence is that it can overestimate the original distribution's uncertainty \cite{minka2005divergence,murphy2012machine}. More details about understanding the over-estimation and under-estimation can be found in the supplemental material.   
The KL divergence is asymmetrical. A variant of the forward KL acts in the opposite direction, and is known as the \textbf{\emph{reverse KL}}: $ KL(p^s||p^t) = \sum_{x} p^s(x) \log\left(\frac{p^s(x)}{p^t(x)}\right)$.
%
Forward KL is known to exhibit \textit{mean-seeking} behavior \cite{minka2005divergence,murphy2012machine}. Looking at its expression, the quantity inside the log is high when $p^t(x)$ is high and $p^s(x)$ is low. To avoid this high value during the minimization process, the mass of $p^s$ should be spread out wherever $p^t(x)$ has some mass, which explains the mean-seeking property. 
Conversely, reverse KL is known to have a \textit{mode-seeking} property \cite{minka2005divergence,murphy2012machine}. To minimize this term, the mass of $p^s$ should cover the region where $p^t(x)$ is high making $p^s$ prioritize covering high modes of $p^t$ and ignoring other modes, which explains the mode-seeking property. 
We provide in the supplemental materials details and examples of understanding the effects of mean-seeking and mode-seeking with respect to the student's uncertainty.
Given these properties of forward and reverse KL, and to ensure better alignment and higher fidelity between student and teacher softmax distributions, in our proposed method, two major changes are made in the original KD loss: (1) instead of using the conventional forward KL in the student's objective function, we use both forward and reverse KL, alongside with a sample-wise weighting mechanism that balances the uncertainty of the student model and (2) reverse KL divergence is used for the teacher's objective function (see Eq.\ref{eq:teacher_obj} and Eq.\ref{eq:student_obj}).
The loss functions of the student and teacher for our approach are shown in Eq. \ref{eq:student_obj} and Eq. \ref{eq:teacher_obj}:
\begin{equation}
\label{eq:student_obj}
L^s = \alpha_s \sum_i L_{CE}(p_{i}^s, y_i) + L_{KD}^s
\end{equation}
where,
\begin{equation}
\label{eq:weighted_eq}
    L_{KD}^s =  \tau^2 \beta_s \sum_i \delta_{i, f} KL(p_{i,\tau}^t||p_{i, \tau}^s)
+ \tau^2 \beta_s \sum_i \delta_{i, r} KL(p_{i, \tau}^s||p_{i,\tau}^t)
\end{equation}
and,
\begin{equation}
\label{eq:teacher_obj}
L^t = \alpha_t \sum_i L_{CE}(p_{i}^t, y_i) + L_{KD}^t 
\end{equation}
where,
\begin{equation}
L_{KD}^t  = \tau^2 \beta_t \sum_i KL(\hat{p}_{i,\tau}^t||\hat{p}_{i, \tau}^s)
\end{equation}
where $p^t$ and $p^s$ are the teacher’s and student’s predictions, respectively and $\tau$ is the temperature to soften these probabilities for distillation. $i$ represents the $i$-th samples in the training set. $y_i$ is the ground truth label of sample $i$, one-hot encoded. 
We see from Eq. \ref{eq:teacher_obj} that the teacher’s KL loss is different than those used in other online methods. In our method, the reverse KL term used in the teacher objective function acts as a regularizer, which combats the overconfidence problem commonly observed in large networks. 

\noindent \underline{\textbf{The Sample-wise Weighting Mechanism.}}
Let's assume $f_s$ and $f_t$ are the representations of the student and the teacher respectively, where the teacher's logit, $z_i^t$, for the $i$-th sample ($x_i, y_i$), is: ${z_i}^t = f_t(x_i,y_i)$. Similarly, we define the student's logit as ${z_i}^s = f_s(x_i,y_i)$. The output of the two networks is a categorical predictive distribution over the $c$ classes: $p^s (y_i=j|x_i)= \sigma(z_i^{s,\tau}) = {\exp({z_i^s}/{\tau})}/{\sum_{j=1}^{c}{\exp({z_j^s}/{\tau})}}$ and $p^t (y_i=j|x_i)= \sigma(z_i^{t,\tau}) = {\exp({z_i^t}/{\tau})}/{\sum_{j=1}^{c}{\exp({z_j^t}/{\tau})}}$. The per-sample entropy of the predictive distribution is given as:
\begin{gather*}
    H_t(x_i, y_i)= - \sum_{j=0}^c p^t(y_i=j|x_i) \log(p^t(y_i=j|x_i))\\
    H_s(x_i, y_i)= - \sum_{j=0}^c p^s(y_i=j|x_i) \log(p^s(y_i=j|x_i)).
\end{gather*}
These entropy values are used to balance the forward and backward KL terms of the student's objective function:
\begin{gather*}
    \delta_{i,r} = \left\{ 
  \begin{array}{ c l }
    1 & \quad \textrm{if } {H_s(x_i, y_i) - H_t(x_i, y_i)} < 0 \\
    v                 & \quad \textrm{otherwise}
  \end{array}
\right. \\
    \delta_{i,f} = \left\{ 
  \begin{array}{ c l }
    v & \quad \textrm{if } {H_s(x_i, y_i) - H_t(x_i, y_i)} < 0 \\
    1                 & \quad \textrm{otherwise}
  \end{array}
\right.
\end{gather*}
where $\delta_i^f$ and $\delta_i^r$ are respectively the per-sample weights of the forward and reserve KL loss in Eq.\ref{eq:student_obj} and $v$ is a hyper-parameter with values greater than one. The term $H_s(x_i, y_i) - H_t(x_i, y_i)$ measures the difference between student and teacher entropies per input sample, with negative values indicating a "mode-seeking" student (underestimation of the teacher’s uncertainties). To address this, we increase weights on forward KL with hyperparameters $\delta_{i,f}$ set to $v >> 1$ and $\delta_{i,r} = 1$. Conversely, positive values signify a "mean-seeking" student with higher entropy (overestimation of the teacher’s uncertainty). In this case, we boost reverse KL weights with $\delta_{i,f} = 1$ and $\delta_{i,r} = v$ for re-adjustments.

We offer a comprehensive and lucid breakdown of our algorithm, presented in a PyTorch-style format in Algorithm 1 of the supplemental material. Our code can easily be integrated into a spectrum of distillation frameworks and across diverse downstream tasks, further enhancing its adaptability and utility. We intend to make the code associated with our paper publicly available upon its acceptance for publication.

\begin{table*}[t]
\caption{Accuracy (\%) comparison on state-of-the art online distillation techniques and the baseline offline vanilla KD technique. (*) are values provided in \cite{qian2022switchable}. Results for CIFAR-10, CIFAR-100, and Tiny-ImageNet are taken across three random seeds. We report the mean and the standard deviation. We observe that BD-KD substantially improves both student and teacher networks. }
\label{tab:sota_final}
\centering
\resizebox{\textwidth}{!}{
\begin{tabular}{lcccccc} \toprule
              & Networks    &   Vanilla KD(*)\cite{hinton2015distilling}    & DML(*)\cite{dml}            & KDCL(*)\cite{kdcl}          & SwitOKD(*)\cite{qian2022switchable}       & BD-KD (Ours) \\ \midrule
        \multicolumn{7}{c}{CIFAR-10} \\ \midrule
        Student & WRN-16-1 & 91.45 $\pm$ 0.06    & 91.96 $\pm$ 0.08  & 91.86 $\pm$ 0.11 & 92.50 $\pm$ 0.17  & \textbf{92.69 $\pm$ 0.18} \\
        Teacher & WRN-16-8 & 95.21 $\pm$ 0.12    & 95.06 $\pm$ 0.05  & \textbf{95.33 $\pm$ 0.17} & 94.76 $\pm$ 0.12 & 94.43 $\pm$ 0.10 \\
        \midrule 
        \multicolumn{7}{c}{CIFAR-100} \\ \midrule
        Student & 0.5MobileNetV2 & 60.07 $\pm$ 0.40 & 66.23 $\pm$ 0.36 & 66.83 $\pm$ 0.05 & 67.24 $\pm$ 0.04 &\textbf{ 68.69 $\pm$ 0.18} \\
        Teacher & WRN-16-2       & 72.90 $\pm$ 0.09 & 73.85 $\pm$ 0.21 & 73.75 $\pm$ 0.26 & 73.90 $\pm$ 0.40 &\textbf{ 74.28 $\pm$ 0.06 }\\
        \midrule
        \multicolumn{7}{c}{Tiny-ImageNet} \\ \midrule
        Student & 1.4MobileNetV2 & 50.98 $\pm$ 0.32 & 55.70 $\pm$ 0.61 & 57.79 $\pm$ 0.30 & 58.71 $\pm$ 0.11 &  \textbf{59.05 $\pm$ 0.17}\\
        Teacher & ResNet34       & 63.18 $\pm$ 0.37 & 64.49 $\pm$ 0.43 & 65.47 $\pm$ 0.32 & 63.31 $\pm$ 0.04 & \textbf{ 66.27 $\pm$ 0.07}\\
        Student & ResNet20       & 52.35 $\pm$ 0.15 & 53.98 $\pm$ 0.26 & 53.74 $\pm$ 0.39 & 55.03 $\pm$ 0.19 &  \textbf{55.19 $\pm$ 0.02 }\\
        Teacher & WRN-16-2       & 56.59 $\pm$ 0.22 & 57.54 $\pm$ 0.19 & 57.71 $\pm$ 0.30 & 57.41 $\pm$ 0.06 &  \textbf{58.29 $\pm$ 0.07 }\\
        \midrule
        \multicolumn{7}{c}{ImageNet Top-1} \\ \midrule
        Student & ResNet18       & 69.76    & 70.81  & 70.91  & \textbf{71.75} & \textbf{71.75} [90.40 top-5]\\
        Teacher & ResNet34       & 73.27    & 73.47  & 73.70  & 73.65 & \textbf{74.02} [91.77 top-5] \\
        \bottomrule
\end{tabular}}
\end{table*}
\begin{table}[h]
\caption{ Accuracy (\%) comparison of student network performances with other KD techniques using similar architectures (SA) or different architecture (DA). Gain(KD) and Gain(DML) represent, respectively, the performance improvement over the classical KD and online KD baseline DML. Most Values are taken from \cite{zhao2022decoupled} and \cite{li2022shake}, and values with (*) are from our own implementation. Results are averaged across 3 different independent runs. We adopted the benchmarking approach outlined in this source \cite{tian2019contrastive}. We put the best performing performance in bold and underlined the second best.}
\label{tab:offlinekd}
\centering
\resizebox{0.6\textwidth}{!}{
\begin{tabular}{ lcccccc} \toprule
        &\multicolumn{3}{c}{\textbf{Same Architecture}} &  \multicolumn{3}{c}{\textbf{Different Architecture}} \\\midrule
        Teacher & WRN-40-2 & R32x4 & VGG13  & VGG13 & R50 & R32x4 \\
        Student & WRN-16-2 & R8x4  & VGG8   & MV2   & VGG8 & ShV2  \\ \midrule
        Teacher & 75.61    & 79.42 & 74.64  & 74.64 & 79.34 & 79.42 \\ 
        Student & 73.26    & 72.50 & 70.36 & 64.60 & 70.36 & 71.82 \\ \midrule
        Vanilla KD
        \cite{hinton2015distilling} & 74.92 & 73.33 & 72.98 & 67.37& 73.81 & 74.45 \\ 
        FitNets
        \cite{romero2014fitnets}    & 73.58  & 73.50 & 71.02 & 64.14 & 70.69 & 73.54 \\
        CRD 
        \cite{tian2019contrastive}      & 75.48 & 75.51 & 73.94 & \textbf{69.73} & 74.30 & 75.65 \\
        DKD 
        \cite{zhao2022decoupled}      & 76.24 & 76.32 & 74.68 & \underline{69.71} & 74.50 * & \textbf{77.07 }\\
        \midrule
        DML
        \cite{dml}      & 75.33   & 74.30 & 73.64 & 68.52 & 74.22& 75.71 *\\
        KDCL
        \cite{kdcl}     & 74.25    & 74.03 & 71.26 & 65.76  & 73.03 & 75.63 *  \\    
        SHAKE
        \cite{li2022shake} & \textbf{76.62}   & \textbf{77.35} &\underline{74.84}  & 70.03& \underline{74.76} & 76.48 *  \\
        CTKD \cite{li2023curriculum} & 75.45 & 74.20 & 73.22 & 68.72& 73.45 *& 75.42 \\
        ER-KD \cite{su2023knowledge} & 75.69 & 75.25 & 74.02 & 68.95& 74.42 *& 75.87 \\
        ER-CTKD \cite{su2023knowledge} & 75.74 & 75.28 & 73.69 & 68.22& 74.25 *& 76.10 \\
        BD-KD (Ours) & \underline{76.47} & \underline{76.67} & \textbf{74.96} & 69.33 & \textbf{74.87} & \underline{76.95} \\
        Gain (KD) & \textcolor{teal}{+ 1.55} &\textcolor{teal}{+ 3.34} & \textcolor{teal} {+ 1.98} & \textcolor{teal}{+ 1.96} & \textcolor{teal}{+ 1.06} & \textcolor{teal}{+ 2.5}     \\
        Gain (DML) & \textcolor{teal}{+ 1.14} & \textcolor{teal}{+ 2.37} & \textcolor{teal}{+ 1.32} & \textcolor{teal}{+ 0.81} & \textcolor{teal}{+ 0.65} & \textcolor{teal}{+ 1.24}\\
        \bottomrule
        
\end{tabular}
}
\end{table}

\section{Experiments and Results}
\subsection{Experimental Details}
In this section, we perform a series of experiments to demonstrate the effectiveness of the proposed online distillation training BD-KD. We compare our method to state-of-the-art approaches using various network architectures and different classification datasets.
\\
\underline{\textit{\textbf{Datasets.}}} We performed our experiments on four different datasets: CIFAR10, CIFAR100 \cite{krizhevsky2009learning}, Tiny-ImageNet \cite{le2015tiny}, and ImageNet \cite{deng2009imagenet}. We use standard data augmentations such as Random crop and Random horizontal flip. We normalized all images channel-wise using the means and standard deviations.
In addition to vanilla KD, we compare our method with recent works on online distillation such as DML \cite{dml}, KDCL \cite{kdcl}, SwitOKD \cite{switokd}, and SHAKE \cite{li2022shake}. We also contrast our findings with current offline KD methods, such as CRD \cite{tian2019contrastive}, DKD \cite{zhao2022decoupled}, FitNet \cite{romero2014fitnets}, etc.
\\
\underline{\textit{\textbf{Training Details.}}} We follow the training procedure of previous works \cite{switokd,tian2019contrastive} and we report the performance on multiple datasets. We mainly use SGD optimizer with 0.9 momentum for all datasets. For CIFAR100, the total number of epochs is set to 240. The initial learning rate is set to 0.01 for most networks, and we perform a scheduler on the learning rate. The latter is divided by 10 every 150th, 180th, and  210th epochs. 
For CIFAR-10, the total number of epochs is set to 300. 
We divide the learning rate by 10 every 100th, 150th, and 200th epoch. For Tiny-ImageNet, the total number of epochs is set to 120 and the initial learning rate is at 0.01. We set the temperature $\tau$ for KD to 2. The chosen value for $v$ is 2 for the adaptive sample-wise weighting. We evaluate the performance of the models using classification accuracy.\\
We align our experiments with the common benchmarking practices within the KD community, ensuring a fair and unbiased evaluation of our method. All teacher-student pairs examined in our paper were sourced from previous papers in which comprehensive hyperparameter tuning had been conducted to achieve optimal results.
\subsection{Results}
\underline{\textit{\textbf{Results on CIFAR-10, 100, and Tiny-ImageNet.}}}
We show the experimental results on CIFAR-10, CIFAR-100, and Tiny ImageNet and compare BD-KD to recent online and offline KD methods. 
Table \ref{tab:sota_final} presents the validation accuracy in \% of recent online KD methods. Overall, our approach gives improvements across all datasets and all student-teacher pairs. With BD-KD, we observe an increase between 1\% - 3\% compared to the online distillation baseline DML \cite{dml} across all datasets and teacher-student pairs. Indeed, a substantial-high increase in student performance with the architecture of mobileNetv2 distilled from WRN-16-2 on CIFAR-100 is observable.
We see a boost of performance by 8\%  when compared to vanilla KD, by almost 2\% when compared to DML and KDCL, and by 1\% compared to SwitOKD. Our method shows a substantial increase in the teacher's performance as well. For example, for Tiny-ImageNet, we witness an increase of almost 3\% in the teacher's performance when compared to SwitOKD. This performance gain on both student and teacher can be explained by the fact that with our method, during training, both networks produce a better quality of transferable knowledge via their logits. Also, complicated training that the teacher endures with techniques like SwitOKD hinders the teacher's accuracy. In addition, our teacher is trained with reverse KL (mode-seeking), which tries to fit the student's noisy distribution in early epochs. This works as a regularization on the teacher, avoiding over-fitting and yielding better performance than current online KD techniques such as DML, KDCL, and SwitOKD.
To further validate our method, we also compare BD-KD with other offline KD techniques. Table \ref{tab:offlinekd} shows the performance on the test set on CIFAR-100 using various pairs of teacher-student architectures. Notably, our method exhibits consistent improvement in all teacher-student pairs over the baseline classical vanilla KD and the baseline online KD (e.g. DML). BD-KD achieves comparable or even better accuracy than other offline or hybrid (switching or combining offline and online KD) KD-based methods. We notice that with BD-KD, the performance gain over the baselines is greatest when the teacher-student pair are of similar architectures. One explanation for this is that with similar architectures, it is easier for both models to update each other's weights and optimize accordingly hence better performance.
\begin{table}[t]

\caption{\footnotesize Top-1 and top-5 accuracy (\%) on the ImageNet validation set. We used ResNet50 to be our teacher network and MobileNetV1 as our student.}
\label{tab:imagenet2}
\centering
\resizebox{\linewidth}{!}{
\begin{tabular}{c|ccccccccc} \toprule
                Acc./Method & Vanilla KD \cite{hinton2015distilling} & CRD \cite{tian2019contrastive} & ReviewKD \cite{chen2021distilling} & DKD \cite{zhao2022decoupled} & AT \cite{zagoruyko2016paying};\cite{mirzadeh2020improved} & RKD \cite{park2019relational} & DML \cite{dml} & SHAKE  \cite{li2022shake}  &  BD-KD (Ours)\\
                \midrule
                Top-1 & 70.68 & 71.40 & 72.56 & 72.05 & 70.72 & 71.32 & 71.13 & 72.66 &\textbf{72.97 } \\
                Top-5 & 90.30 & 90.42 & 91.00 & 91.05 & 90.03 & 90.62 & 90.22 & 91.35 & 90.98 \\
                \bottomrule
        
\end{tabular}}
\end{table}
%
\\
\underline{\textit{\textbf{Results on ImageNet.}}}
Top-1 and top-5 accuracies on ImageNet are reported in Table \ref{tab:sota_final}  and Table \ref{tab:imagenet2} for teacher-student pairs ResNet18 - Resnet32 and Resnet50 - MobilenetV1. Our proposed method BD-KD produces substantially higher or similar performance when compared to the baseline KD methods whether online or offline and as well as other KD techniques. For instance, we achieve a performance gain of 2.29 \% when compared to the vanilla KD and 1.84 \% compared to DML.
We even reached an approximately 0.31\% increase when compared to recent hybrid KD methods such as SHAKE \cite{li2022shake}, which use both offline and online KD through a proxy teacher setting. This shows that our technique is scalable to large-scale datasets.
\\
\noindent \underline{\textit{\textbf{Mitigating the capacity gap problem.}}}
\begin{figure}[tb]
  \centering
  \begin{subfigure}{0.32\linewidth}
  \includegraphics[width=1.2\textwidth]{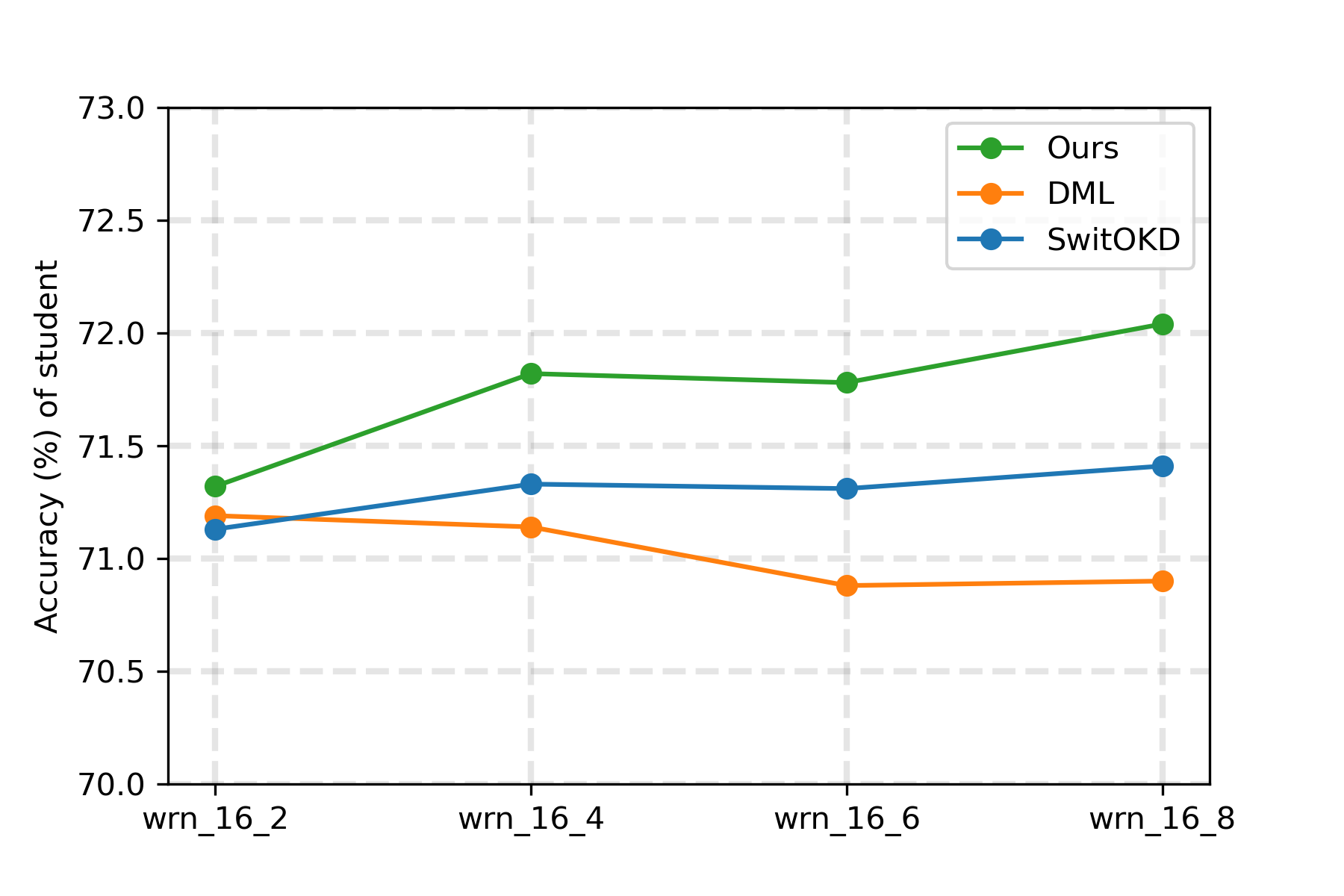}
    \caption{}
    \label{fig:capacity_short_a}
  \end{subfigure}
  \hfill
  \begin{subfigure}{0.3\linewidth}
  \includegraphics[width=\textwidth]{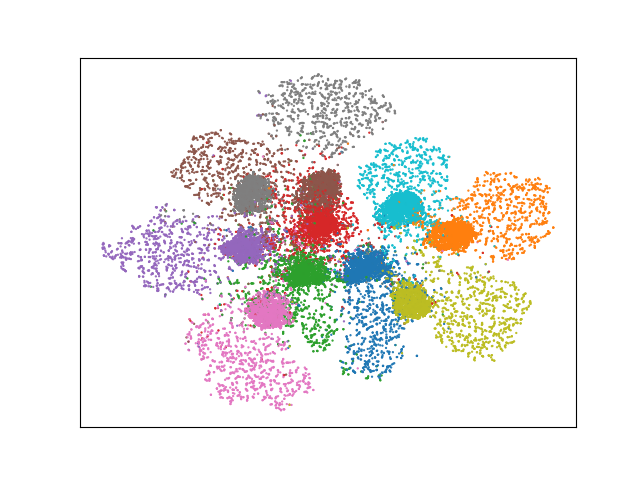}
    \caption{}
    \label{fig:tsne_switch}
  \end{subfigure}
  \hfill
  \begin{subfigure}{0.3\linewidth}
  \includegraphics[width=\textwidth]{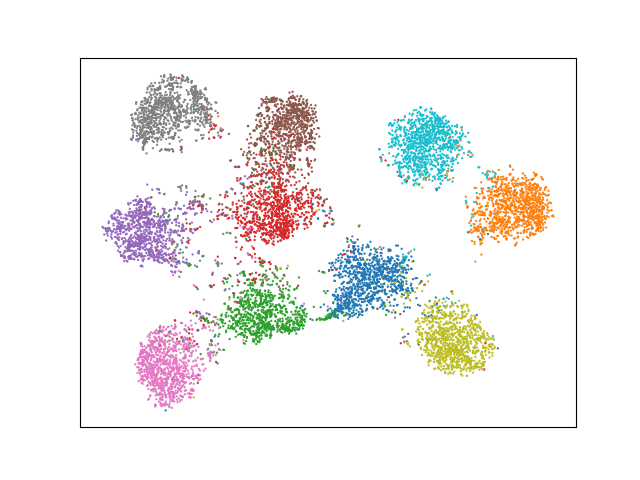}
    \caption{}
    \label{fig:tsne_bdkd}
  \end{subfigure}
  \caption{(a) \textbf{Capacity gap Curve}; Student (ResNet20) distilled from different teacher capacity (WRN-16-2 to WRN-16-8) on CIFAR100. (b) \textbf{TSNE visualization} \cite{van2008visualizing} of the penultimate feature layer of the student model (WRN-16-1) trained with SwitOKD on the test images from CIFAR10. (c) \textbf{TSNE visualization }\cite{van2008visualizing} of the penultimate feature layer of the student model (WRN-16-1) trained with BD-KD on the test images from CIFAR10. }
  \label{fig:tsne_capcity}
\end{figure}

\noindent One of the main consequences of the capacity gap problem in KD is that the student does not benefit from larger teacher distillation \cite{mirzadeh2020improved}. In Figure \ref{fig:capacity_short_a}, we compare the available online distillation techniques and observe their behavior with respect to the capacity gap problem. Predominantly, BD-KD is performing well overall. We observe a steady boost as we increase our network size. As a matter of fact, with DML, online distillation has reduced performance when a higher-capacity teacher is used. Similarly, with SwitOKD, student performance plateaus as we increase the teacher  size.

\begin{table*}[t]
\caption{Accuracy with three network setting using online distillation techniques on CIFAR-100 dataset. (*) are values provided in \cite{qian2022switchable}. The network with (T) or (S) means it serves as a teacher or a student , respectively. (S/T) means that the network serves as both a teacher and a student at the same time.}
\centering
\resizebox{\textwidth}{!}{
\begin{tabular}{llllllll}
\toprule
\multicolumn{1}{c}{Networks} & \multicolumn{1}{c}{Vanilla(*)} & \multicolumn{1}{c}{DML(*)} & \multicolumn{1}{c}{KDCL(*)} & \multicolumn{1}{c}{\begin{tabular}[c]{@{}c@{}}SwitOKD(*)\\ (1T2S)\end{tabular}} & \multicolumn{1}{c}{\begin{tabular}[c]{@{}c@{}}SwitOKD(*)\\ (2T1S)\end{tabular}} & \multicolumn{1}{c}{\begin{tabular}[c]{@{}c@{}}BD-KD (Ours) \\ (1T2S)\end{tabular}} & \multicolumn{1}{c}{\begin{tabular}[c]{@{}c@{}}BD-KD(Ours)\\ (2T1S)\end{tabular}} \\ \hline
MobileNet                    & 58.65  (S)                     & 63.75 (S)                  & 62.13 (S)                   & 64.62    (S)                                                                    & 65.03     (S)                                                                   & \textbf{67.72 } (S)                                                                   & \textbf{68.36} (S)                                                                    \\ 
WRN-16-2                     & 73.37   (S/T)                    & 74.30   (S/T)                & 73.94  (S/T)                  & \textbf{75.02} (S)                                                                        & 71.73 (T)                                                                         & 74.70 (S/T)                                                                     & 74.62    (S/T)                                                                 \\ 
WRN-16-10                    & 79.45 (T)                       & 77.82  (T)                 & \textbf{80.71}  (T)                  & 77.33   (T)                                                                     & 77.07 (T)                                                                       & 79.26   (T)                                                                  & 79.05           (T)                                                          \\ \bottomrule
\end{tabular}
}
\vspace{-4mm}
\label{tab:extension}
\end{table*}
\noindent \underline{\textit{\textbf{Extension to multiple networks.}}}  Similar to DML and SwitOKD, BD-KD can be extended for multiple networks. Since the loss functions between the teacher and the student are different, we adopt the same multiple-teacher setting as SwitchOKD. For instance, in a three-network training, there are two approaches to training the student model: the one-teacher-two-student setting (1T2S) and the two-teacher-one-student setting (2T1S). As can be seen in Table \ref{tab:extension}, we achieve a higher accuracy student across all methods using the 2T1S scenario, whereas the student for 1T2S is a little lower. An explanation for this is that for 2T1S, the compact student learns an ensemble-like and diverse knowledge from multiple teachers. As for our teacher networks, our largest teacher with BD-KD achieves comparable results to the vanilla KD at 79.45\% and KDCL at 80.71\% but much higher than DML and SwitOKD 1T2S and 2T1S. More details on these multi-network settings are provided in the supplemental material.
%

\begin{table*}[t]
\caption{ \textbf{Top table:} Comparison of model calibration performances in ECE of student networks with vanilla offline KD and recent online distillation techniques. For ECE, the lower the better. \textbf{Bottom table:} Expected Calibration Error (ECE) of Student MobileNet and WRN-16-2  trained with WRN-16-10 teacher on CIFAR-100. Results show that BD-KD achieves lower ECE overall.}
\label{tab:ece}
\centering
\resizebox{0.6\textwidth}{!}{
\begin{tabular}{ lccccccc} \toprule
        Teacher & WRN-40-2 & ResNet32x4 & VGG13   & ResNet50  & WRN-16-6 \\
        Student & WRN-16-2 & ResNet8x4  & VGG8      & VGG8  & ResNet20 \\ \midrule
        Vanilla KD & 6.45\% & 5.81\% & 8.46\% & 7.21\%  & 11.93\% \\
        DML      & 3.33\% & 2.60\% & 7.72\%  & 4.81\% & 4.79\% \\
        SwitOKD & 4.36\% & \textbf{2.01\%} & 6.12\%  & 6.49\&  & 4.65\% \\
        BD-KD (Ours) & \textbf{3.15\%} & 2.59\% & \textbf{4.03\%}  & \textbf{4.27\%}   & \textbf{2.40\%} \\
        \bottomrule
\end{tabular}}
\vspace{3mm}
\vfill
\resizebox{0.6\textwidth}{!}{
\begin{tabular}{lcc}
\toprule
\textbf{Networks}                                 & \textbf{SwitOKD} & \textbf{BD-KD (ours)} \\ \midrule
MobileNet (trained with 1T2S)                     & 3.4\%            & 2.8\%                 \\ \midrule
\multicolumn{1}{l}{WRN-16-2 (trained with 1T2S)} & 3.7\%            & 3.4\%                 \\ \bottomrule
\end{tabular}}
\vspace{-2.5mm}
\end{table*}

\begin{figure}[!h]
\scalebox{0.6}{
\centering{
    \noindent\begin{minipage}{0.4\textwidth}
    \includegraphics[width=\linewidth]{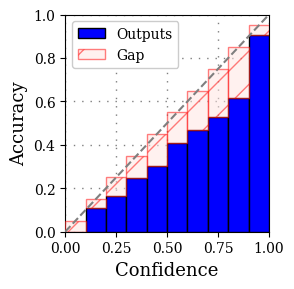}
    \centering{(a) Vanilla-KD \\ ECE =0.1207 }
    \end{minipage}
    \hfill
    \begin{minipage}{0.4\textwidth}
    \includegraphics[width=\linewidth]{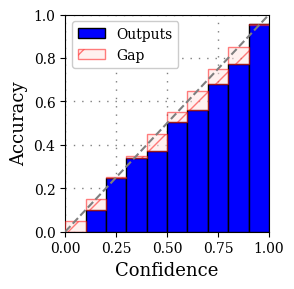}
    \centering{(b) DML \\ ECE =0.04298 }
    \end{minipage}\hfill
    \begin{minipage}{0.4\textwidth}
    \includegraphics[width=\linewidth]{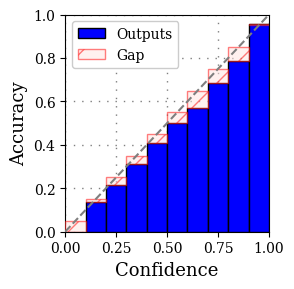}    
    \centering{(c) SwitOKD \\ ECE = 0.04108}
    \end{minipage}
    \hfill
    \begin{minipage}{0.4\textwidth}
    \includegraphics[width=\linewidth]{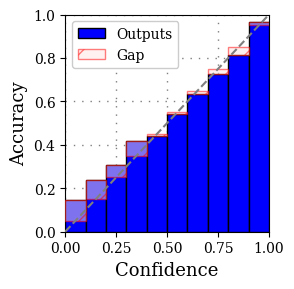}
    \centering{(d) BD-KD (Ours) \\  ECE = 0.0252}
    \end{minipage}
    }}
    \centering \caption{\textbf{Calibration curves} (ResNet 20 student, teacher WRN-16-8). Online KD methods improve calibration compared to vanilla KD, and of the tested online methods BD-KD improves calibration the most.}
    \label{fig:calibration}
\end{figure}

\noindent \underline{\textbf{\textit{Calibration Analyses.}}} To quantify and measure model calibration performance, we use the Expected Calibration Error \cite{naeini2015obtaining}. We also qualitatively evaluate calibration with reliability curves (also called calibration curves). Figure \ref{fig:calibration} shows the calibration curves of a student network ResNet20 distilled from a WideResNet of depth 16 and width 8 (WRN-16-8) using offline vanilla KD and online KD methods on CIFAR100. We observe that performing offline distillation gives an over-confident compact student network as compared to online KD techniques. With offline vanilla KD and by making the student match the confidence of the large pre-trained network, the confidence of the compact network is easily affected, shifting it to be more over-confident. 

\noindent A solution to mitigate miscalibration during distillation is to opt for online KD techniques. In Figure \ref{fig:calibration} (b), (c), and (d) we see that the same compact online student network is better calibrated than the baseline offline. We see a drop in the ECE from 0.1207 for offline to 0.0429 with DML, 0.04108 with SwitOKD, and an even lower drop to 0.0252 using BD-KD.\\
\noindent Table \ref{tab:ece} (top table) reports the student expected calibration error ECE on the test set of CIFAR-100 with different teacher-student pairs and various KD methods. We observe that BD-KD produces lower values of ECE, meaning that the resulting compact student is better calibrated. Indeed, we accomplish an ECE reduction by almost half when distilling WRN-16-6 to ResNet20 and similarly when distilling VGG13-VGG8 with BD-KD compared to DML and SwitOKD. This decrease in the ECE metric indicates that the proposed weighting scheme related to balancing the forward and backward KL during the student's training to overcome the over-confidence and under-confidence of the student model helps in calibrating the student network. Additionally, the obtained student network has more compact class clusters with better class separation (in Figure \ref{fig:tsne_bdkd}) than SwitOKD (in Figure \ref{fig:tsne_switch}), which makes for better generalization and hence better calibration. In some cases, BD-KD showed almost similar performance as SwitOKD. We argue that the optimization process of SwitOKD is sophisticated, which may lead sometimes to different outcomes of calibration results depending on the teacher network and the initialization of both networks. 
\noindent Table \ref{tab:ece} (bottom table) shows the ECE of student networks MobileNet and WRN-16-2 trained simultaneously with one single teacher network WRN-16-10 with SwitOKD and BD-KD. 
We observe that our method resulted in a lower ECE error overall (the lower the better). We observe a higher drop in ECE with the smallest capacity student network using our method.
\\
\begin{table}[ht]
    \centering
    \caption{Accuracy and ECE performances of both student (WRN-16-2) and teacher (WRN-40-2) networks under different KD setting on CIFAR-100. }
    \label{tab:loss_ablations}
\centering
\begin{tabular}{lccc}
\toprule
KD strategy&  \multicolumn{2}{c}{Accuracy
T/S}& ECE\\
\midrule
 Vanilla& 75.84& 75.06& 6.45\%\\
 Vanilla + BD-KD EQ2& 75.84&  75.43& 4.02\%\\
 Vanilla  + JS& 75.84& 75.34&5.1\%\\
 \midrule
 DML& 78.57& 75.81&3.60 \%\\
 DML + BD-KD EQ2& 78.29&  75.89 & 3.20 \%\\
 DML + JS& 78.35& 75.82&3.51\%\\
 \midrule
 SwitOKD& 75.84& 74.90&4.40 \%\\
 SwitOKD + BD-KD EQ2& 77.10&  75.93& 3.21 \%\\
 \midrule
 \midrule
 BD-KD (OURS)& 78.44& \textbf{76.26}& \textbf{3.00\%}\\
 \bottomrule
\end{tabular}

\end{table}

\subsection{Analyses and Ablations}
\begin{figure}[t] 
   \begin{subfigure}{0.35\textwidth}
       \includegraphics[width=\linewidth]{Figures/cifar100_accuracy_gap.png}
       \caption{}
       \label{fig:subim1}
   \end{subfigure}
\hfill 
   \begin{subfigure}{0.35\textwidth}
       \includegraphics[width=\linewidth]{Figures/imagenet_accuracygap_training.png}
       \caption{}
       \label{fig:subim2}
   \end{subfigure}
\hfill 
   \begin{subfigure}{0.25\textwidth}
       \includegraphics[width=\linewidth]{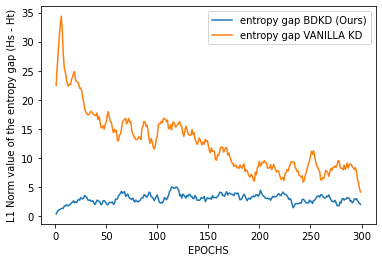}
       \caption{}
       \label{fig:subim3}
   \end{subfigure}

   \caption{(a) Accuracy gap between teacher (ResNet32x4) and student (ShuffleNetV2) networks on the test set of CIFAR-100 dataset using BD-KD and SwitOKD.(b) Accuracy gap between teacher (ResNet50) and student (MBV1) networks on the validation set of the ImageNet dataset using BD-KD and SwitOKD.  (c) Evolution of $H_t$ and $H_s$ during training of student (MobileNet V2) and teacher (WRN-16-2) on CIFAR100.}
   \label{fig:all_gaps}
\end{figure}

\begin{figure}[tb]
  \centering
  \begin{subfigure}{0.3\textwidth}
  \includegraphics[width=\textwidth]{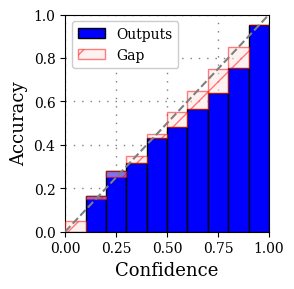}
    \caption{}
    \label{fig:ece1}
  \end{subfigure}
  \centering
  \begin{subfigure}{0.3\textwidth}
  \includegraphics[width=\textwidth]{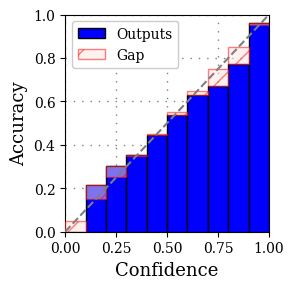}
    \caption{}
    \label{fig:ece2}
  \end{subfigure}
\caption{Student WRN-16-2 distilled from teacher WRN-40-2 on CIFAR100. (a) Student network trained without weighting (i.e. $\delta_1 = 1$ and $\delta_2 = 1$. (b) Student network trained using our proposed sample-weighting of the KL terms according to Eq. \ref{eq:teacher_obj} and Eq.\ref{eq:student_obj}.}
  \label{fig:ablation_calibration}
\end{figure}

\underline{\textit{\textbf{BD-KD for offline KD settings?}}} Consistent with earlier discussions, BD-KD seamlessly integrates into various KD losses, both in offline and online KD configurations. In Table \ref{tab:loss_ablations}, we provide a comparative analysis of the performance of student and teacher models trained under the offline KD setting, also known as vanilla KD. Specifically, in offline KD, we replace the conventional student loss with that of BD-KD (Eq. \ref{eq:weighted_eq}). Our findings reveal a notable enhancement in performance with Vanilla KD + BD-KD (75.43\% accuracy) compared to vanilla KD alone (75.06\% accuracy).\\
\noindent \underline{\textit{\textbf{Utilizing BD-KD across diverse online KD configurations. }}} We present comparisons in Table \ref{tab:loss_ablations} between DML alone and DML enriched with the student-centered loss of BD-KD (Eq. \ref{eq:weighted_eq}), as well as SwitOKD trained with BD-KD's student loss. We observe consistent performance enhancements, including improvements in accuracy and ECE, across student models trained with our approach. For example, with SwitOKD, we witness an increase from 74.90\% to 75.93\% in accuracy, and a reduction in ECE from 4.40\% to 3.21\%. Thus, BD-KD effectively improves both the performance and calibration of the compact student model.\\
\underline{\textit{\textbf{JS Divergence vs. Sample-wise Reweighting in BD-KD.}}} In Table \ref{tab:loss_ablations}, we contrast various KD settings trained with JS divergence and BD-KD. While JS divergence improves the baselines, BD-KD demonstrates superior performance overall, in terms of both accuracy and model calibration. This highlights the significant advantage of sample-wise weighting across different divergence elements.
Additionally, to have a better understanding of the effect of the sample-wise weighting of both KL terms with respect to model confidence calibration, we compare the calibration curves and the ECE values of a student model trained with sample weighting to that without weighting of the KL divergences in Eq.\ref{eq:student_obj}.
Figure \ref{fig:ablation_calibration} shows that the student model trained without the proposed sample-wise weighting scheme for the KL divergences is miscalibrated with an ECE of 0.0461, whereas BD-KD shows a better-calibrated network with an ECE of 0.0315 (around 1.5\% decrease). 
The reliability diagram of BD-KD shows that our student became under-confident at low confidence values, but almost perfectly calibrated on the important mid-to-high confidence values.\\
\underline{\textit{\textbf{Accuracy and Entropy Gap.}}} We provide in Figure \ref{fig:all_gaps} (b) the accuracy gap between teacher (ResNet50) and student (MobileNetV1) on the ImageNet validation dataset on both BD-KD and SwitOKD. Similarly, in Figure \ref{fig:all_gaps} (a), we present the accuracy gap between the teacher (ResNet32x4) and student (ShuffleNetV2) on CIFAR-100 using BD-KD and SwitOKD. We observe that, throughout the training process, BD-KD can maintain a low training accuracy gap between the teacher and student networks, which contributes to the fidelity factor and hence better generalization and calibration. Additionally, this explains why BD-KD is a student-centered training. By maintaining this gap very low, we make sure that the student network is able to catch up with the teacher's knowledge and confidence. We also observe a lower accuracy gap when using our proposed method, compared to SwitOKD. BD-KD is able to maintain this performance throughout the training process. Furthermore, we observe a decreasing entropy gap in Figure \ref{fig:all_gaps} (c) throughout the learning process of the networks. Indeed, BD-KD is designed to maintain a lower uncertainty gap, enhancing student model calibration, as seen in our experiments.\\
\noindent \underline{\textit{\textbf{Sensitivity to $v$.}}}
\noindent We vary the hyperparameter $v$ according to the following values [1,2,3,4] (in Table \ref{tab:all_sensitivity}) and record the accuracies of both teacher and student networks on the test set of CIFAR-100. We see that the optimal choice for $v$ is 2. As we increase the value of this hyperparameter performance degrades. One must be careful in choosing the appropriate value of $v$ based on the downstream task and the architecture of the student since a high value can lead to unnormalized gradients and even can cause gradient explosion when $v$ is too high.
\\

\noindent \underline{\textit{\textbf{Sensitivity to $\tau$.}}}
We performed hyperparameter sensitivity analysis on $\tau$, varied its values and observed the accuracies in \% of both student and teacher networks in Table \ref{tab:all_sensitivity}. We find that $\tau=2$ is the optimal value. As we increase $\tau$, we observe a decrease in performance in both networks. This could be linked to the fact that higher temperatures contribute to maximizing the entropy gap between the teacher and the student, hence making it very difficult for the student to follow the teacher. 
%
\begin{table}[ht]
    \centering
    \caption{Sensitivity of BD-KD to $v$ and $\tau$}
    \label{tab:all_sensitivity}
        \begin{minipage}[t]{0.35\textwidth}
       \resizebox{\textwidth}{!}{
        \begin{tabular}{lllll}
\toprule
             & \multicolumn{1}{c}{v=1} & \multicolumn{1}{c}{v=2} & \multicolumn{1}{c}{v=3} & \multicolumn{1}{c}{v=4} \\ \midrule
ResNet20 (S) & 71.90                   &  \textbf{72.04}                  & 71.70                   & 71.72                   \\
WRN-40-2 (T) & 77.19                   &  77.31                  & 77.58                   & 77.83                    \\ \bottomrule
\end{tabular}
}
    \end{minipage}%
    \hspace{3mm}
    \begin{minipage}[t]{0.4\textwidth}
        \resizebox{\textwidth}{!}{
        \begin{tabular}{lccccc}
\toprule
        \multicolumn{2}{c}{Networks} & $\tau=1$   & $\tau=2$   & $\tau=3$ & $\tau=4$ \\ \midrule
Student & WRN\_16\_2  & 75.51 & \textbf{76.42} & 76.37 & 76.30 \\
Teacher & WRN\_16\_40 & 78.00 & 78.80 & 79.02  & 78.92 \\ \bottomrule
\end{tabular}
}
    \end{minipage}%
    \hfill
\end{table}

\section{Conclusion}
Our proposed approach addresses the challenges of model compression and network reliability in tandem. By dynamically balancing (i.e. sample-wise weighting) the forward and reverse KL in the student's objective function and using a reverse KL for teacher distillation, we enhance the student’s accuracy and calibration surpassing other online distillation techniques. Extensive experiments validate the effectiveness of our method.
Mainly, our method is loss and task-agnostic, achieves better accuracy and model-calibration performance, making it easily integrated into other objective functions and onto different tasks.

\bibliographystyle{splncs04}
\bibliography{main}

\appendix
\section{Algorithm of the Proposed Method}
\begin{algorithm}[ht]
\SetAlgoLined
\footnotesize
    \PyComment{x:input image} \\
    \PyComment{y:ground truth} \\
    \PyComment{model\_s: student model} \\
    \PyComment{model\_t: teacher model} \\
    \PyComment{H\_s, H\_t: entropy student and teacher} \\
    \PyComment{$v$: hyper-parameter related to BD-KD} \\
    \PyComment{alpha\_s,alpha\_t,beta\_s,beta\_t: hyper parameters related to KD} \\
    \PyCode{for (x,y) in Batches:} \\
    \Indp   
        \PyComment{values of logits} \\
        \PyCode{output\_t = model\_t(x)} \\ 
        \PyCode{output\_s = model\_s(x)} \\ 
        \PyComment{calculate the ce losses} \\
        \PyCode{loss\_ce\_t = beta\_t * ce\_loss(output\_t, y)}\\
        \PyCode{loss\_ce\_s = beta\_s * ce\_loss(output\_s, y)}\\
        \PyComment{calculate the kl loss teacher} \\
        \PyCode{loss\_kl\_t = T*T*alpha\_t * kl\_div\_loss(output\_t, output\_s.detach())}\\
        \PyComment{teacher total loss} \\
        \PyCode{total\_loss\_t= loss\_ce\_t + loss\_kl\_t }\\
        \PyComment{calculate entropy gap} \\
        \PyCode{H\_gap = H\_s - H\_t}\\
        \PyComment{initializing sample weights} \\
        \PyCode{delta\_reverse = 1}\\
        \PyCode{delta\_forward = 1}\\
        \PyCode{if H\_gap $\geq$ 0: }\\
        \Indp
            \PyCode{delta\_reverse = $v$}\\
        \Indm 
        \PyCode{else: }\\
         \Indp
            \PyCode{delta\_forward = $v$}\\
        \Indm 
        \PyComment{calculate the kl loss student} \\
        \PyCode{loss\_kl\_s = T*T*alpha\_s * \\weighted\_kl\_div\_loss(output\_s,output\_t.detach(), delta\_reverse) + \\ weighted\_kl\_div\_loss(output\_t.detach(),output\_s, delta\_forward)}\\
        \PyComment{student total loss} \\
         \PyCode{total\_loss\_s = loss\_ce\_t + loss\_kl\_s}\\
        \PyComment{updates} \\
        \PyCode{total\_loss\_t.backward()}\\
        \PyCode{total\_loss\_s.backward()}\\
    \Indm 
\caption{PyTorch-style pseudocode for BD-KD}
\label{algo:your-algo}
\end{algorithm}
Illustrated in Algorithm 1, the pseudocode of our BD-KD framework bears resemblances to online knowledge distillation, with a pivotal divergence being the infusion of a balancing mechanism via uncertainty (entropy).
Our code can easily be integrated into a spectrum of distillation frameworks and across diverse downstream tasks, further enhancing its adaptability and utility.

\section{Understanding Over-estimation and Under-estimation of Uncertainty}
\begin{figure}[h]
\begin{center}
\scalebox{0.9}{\includegraphics[width=\linewidth]{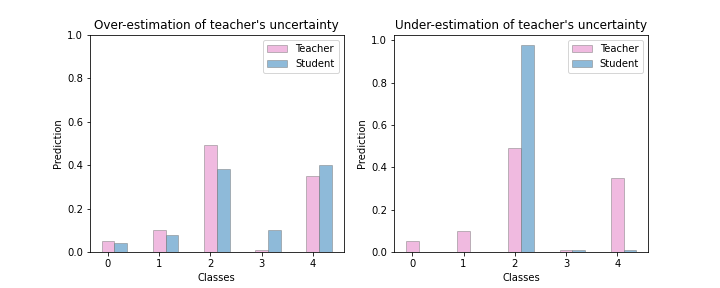}}
\end{center}
\vspace{-0.5cm}
\caption{
Understanding over-estimation (left) and under-estimation (right) of uncertainty. We say that the compact student network under-estimates the uncertainty of the teacher when its prediction entropy is small compared to the teacher's uncertainty.  Similarly, the student over-estimates the teacher's uncertainty when its entropy prediction is higher than the teacher's entropy.   
}
\label{fig:over-under}
\end{figure}

According to Figure 1, When we consider a teacher's softmax distribution over the different classes $p^t$ (in pink), we can observe two distinct outcomes for the student’s distribution $p^s$. The outcome hinges on the choice between using forward or reverse KL divergence. Notably, these choices lead to different estimations of uncertainty. When, employing forward KL, the student's distribution $p^s$ (in blue) is driven to assign non-zero probabilities across all classes. In contrast, reverse KL can result in a $p^s$ that primarily encompasses the dominant mode in $p^t$, allocating the highest probability to a single class while minimizing probabilities for others. This observation highlights that minimizing forward KL might generate distributions with greater uncertainty than the original $p^t$, while minimizing reverse KL could yield the opposite effect.

\section{Experimental Setup }
In this section, we include more details of our experimental setting in our paper. For BD-KD experiments, we perform a hyperparameter search and we report the value that leads to the highest performance. 
In Table \ref{tab:hpt} below, we provide particulars of the hyperparameter space considered per dataset type: CIFAR-10, CIFAR-100, TinyImageNet, and ImageNet. 
\begin{table}[h]
\caption{Hyperparameter search space on CIFAR-10, CIFAR-100, TinyImageNet, and ImageNet. \footnotesize{BS: batch size, LR: learning rate, WD: weight decay}}
\label{tab:hpt}
\centering
\resizebox{\linewidth}{!}{%
\begin{tabular}{lc}
\toprule
Dataset & Hyperparameter search space \\
\midrule
CIFAR-10 and CIFAR-100 & BS=\{64,128\} \\
                    & LR=\{0.1,0.01, 0.02, 0.05\} \\
                    & $\tau$ =\{1,2,3,4\} \\
                    & $v$ = \{2,3,4\}\\
                    & WD = \{1e-4, 5e-4\}\\
                    &$\alpha_S$,$\alpha_T$,$\beta_S$,$\beta_T$ = \{1\}\\
\midrule
TinyImageNet & BS=\{64,128,256\} \\
                & LR=\{0.01, 0.02, 0.05\} \\
                & $\tau$ =\{2\} \\
                & $v$ = \{2\}\\
                & WD = \{1e-4, 5e-4\}\\
                & $\alpha_S$,$\alpha_T$,$\beta_S$,$\beta_T$ = \{1\}\\
\midrule
ImageNet    & BS=\{256, 512\} \\
            & LR=\{0.1, 0.2\} \\
            & $\tau$ =\{1,2\} \\
            & $v$ = \{2\}\\
            & WD = \{1e-4, 5e-4\}\\
            & $\alpha_S$,$\alpha_T$,$\beta_S$,$\beta_T$ = \{1\}\\
            
\bottomrule
\end{tabular}
}
\end{table}

\section{Expected Calibration Error and Reliability Curves}
In order to evaluate the degree of miscalibration of the student network, we use Calibration Curves and Expected Calibration Error measures.

\textbf{Calibration Curves.} To assess if a model is well calibrated, we often plot the \emph{calibration curve}, also referred to as \emph{reliability diagram} \cite{guo2017calibration}. The diagrams are one way of depicting model calibration where the expected accuracy of the network is plotted as a function of its confidence (i.e. softmax probabilities). A "perfectly" calibrated model is represented by the identity function. A deviation above (under-confidence) or below (over-confidenc) the diagonal is an indication of a miscalibrated model. To obtain this curve, first, we partition the confidence estimates into $M=10$ equal interval bins $I_m$. Then, we proceed to calculate the accuracy of each bin as follows: 
\begin{equation*}
    acc(B_m) = \frac{1}{|{B_m}|}\sum_{i \in B_m} \mathbbm{1}(p_i=y_i)
\end{equation*}
where $\hat{y_i}$ and $y_i$ are, respectively, the predicted and true class labels of sample $i$, and $B_m$ is the set of sample indices whose confidence falls into interval $m$.
Finally, we compute the average confidence within each bin: 
\begin{equation*}
    B_m: conf(B_m) = \frac{1}{|{B_m}|} \sum_{i \in B_m}\hat{p}_i
\end{equation*}
where $\hat{p}_i$ is the confidence for sample $i$.

\textbf{Expected Calibration Error.}
In contrast to the calibration curves, which are visual diagrams capturing the degree of model calibration, the Expected Calibration Error (ECE) is a  function that returns a scalar value measuring the degree of miscalibration of the model. 
Similar to the calibration curves, confidence values are partitioned into $M=10$ equal bins. ECE is defined as the weighted average of the difference between the accuracy and confidence at each bin: 
\begin{equation*}
    ECE = \sum_{m=1}^{M}\frac{|{B_m}|}{n} |{acc(B_m) - conf(B_m)}|
\end{equation*}
 

\section{Other Related Calibration Results}


\begin{figure}[h!]
    \begin{minipage}{0.23\textwidth}
    \includegraphics[width=\linewidth]{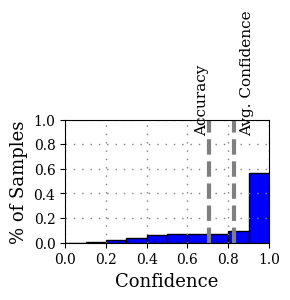}
    \centering{(a) Vanilla-KD }
    \end{minipage}
    \begin{minipage}{0.23\textwidth}
    \includegraphics[width=\linewidth]{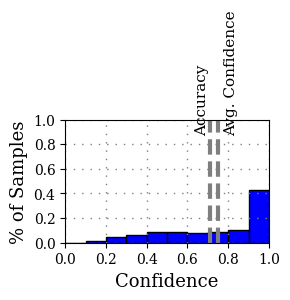}
    \centering{(b) DML }
    \end{minipage}
    \begin{minipage}{0.23\textwidth}
    \includegraphics[width=\linewidth]{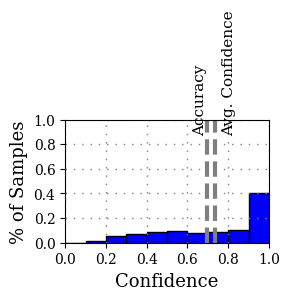}    
    \centering{(c) SwitOKD}
    \end{minipage}
    \begin{minipage}{0.23\textwidth}
    \includegraphics[width=\linewidth]{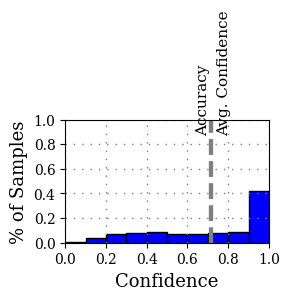}
    \centering{(d) BD-KD (Ours)}
    \end{minipage}
\caption{\textbf{Confidence histograms} (ResNet 20 student, teacher WRN\_16\_8) on CIFAR-100.}
\label{fig:sup1}
\end{figure}
\begin{figure}[h]
    \begin{minipage}{0.23\textwidth}
    \includegraphics[width=\linewidth]{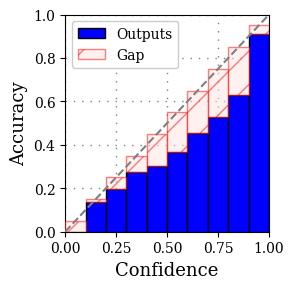}
    \centering{(a) Vanilla-KD \\ ECE =0.119}
    \end{minipage}
    \hfill
    \begin{minipage}{0.23\textwidth}
    \includegraphics[width=\linewidth]{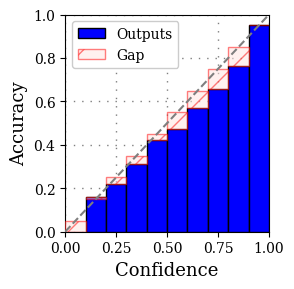}
    \centering{(b) DML \\ ECE =0.0479 }
    \end{minipage}
    \hfill
    \begin{minipage}{0.23\textwidth}
    \includegraphics[width=\linewidth]{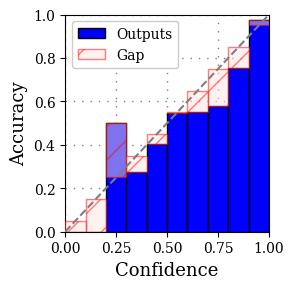}    
    \centering{(c) SwitOKD \\ ECE = 0.0465}
    \end{minipage}
    \hfill
    \begin{minipage}{0.23\textwidth}
    \includegraphics[width=\linewidth]{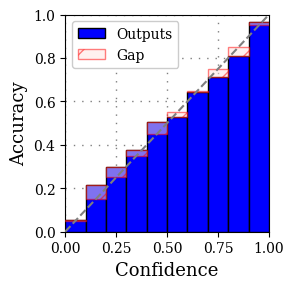}
    \centering{(d) BD-KD (Ours) \\  ECE = 0.0240}
    \end{minipage}
    \caption{\textbf{Calibration curves.} (ResNet 20 student, teacher WRN\_16\_6) on CIFAR-100. Online KD methods improve calibration. Out of  the online methods (DML, SwitOKD), BD-KD improves calibration the most.}
    \label{fig:sup2}
\end{figure}

Figure \ref{fig:sup1} shows the confidence calibration (i.e. the distribution of prediction confidence or probabilities associated with the target labels)  of a ResNet20 student network trained with various KD techniques (vanilla offline KD (a), DML (b), SwitOKD (c ), and BD-KD (d) ). 
With the vanilla offline KD, we observe a lot of concentration in the highest confidence bins (between 0.9- 1.0) and there is a huge gap between the accuracy and the expected confidence when compared to the online KD techniques. For both DML and SwitOKD, we see a slight decrease in the gap between the accuracy of the student model and the average accuracy. We witness that BD-KD (ours) resulted in the most significant decrease in the gap. Indeed, we see that the average confidence of the student ResNet20 matches very closely the expected confidence. 
Figure \ref{fig:sup2}, shows the calibration curves with ResNet20 student with a smaller capacity teacher WRN-16-6 (than WRN-16-10). Similar to our previous observations, BD-KD leads to a better-calibrated model overall (the bars are almost aligned along the diagonal).

\section{Extension to multi-networks (details)}
We present the objective functions for both teacher and student(s) under multiple network settings. $p_{i,\tau}^t$ and $p_{i,\tau}^s$ are the softmax probabilities of the teacher and the student, respectively.
\subsection{One Teacher Two Students (1T2S) loss functions} 
The student losses in this case are:
\begin{multline*}
L_{KD}^{s_1} = \alpha_{s_1} \sum_i L_{CE}(p_{i}^{s_1}, y_i)\\
+ \tau^2 \beta_{s_1} \sum_i {\delta_{i, f}}^{s_1,t} KL(p_{i,\tau}^t||p_{i, \tau}^{s_1})\\
+ \tau^2 \beta_{s_1} \sum_i {\delta_{i, r}}^{s_1,t} KL(p_{i, \tau}^{s_1}||p_{i,\tau}^t)\\
+ \tau^2 \beta_{s_1} \sum_i {\delta_{i, f}}^{s_1,s_2} KL(p_{i,\tau}^{s_2}||p_{i, \tau}^{s_1})\\
+ \tau^2 \beta_{s_1} \sum_i {\delta_{i, r}}^{s_1,s_2} KL(p_{i, \tau}^{s_1}||p_{i,\tau}^{s_2})\\
\end{multline*}
\\
\begin{multline*}
L_{KD}^{s_2} = \alpha_{s_2} \sum_i L_{CE}(p_{i}^{s_2}, y_i)\\
+ \tau^2 \beta_{s_2} \sum_i {\delta_{i, f}}^{s_2,t} KL(p_{i,\tau}^t||p_{i, \tau}^{s_2})\\
+ \tau^2 \beta_{s_2} \sum_i {\delta_{i, r}}^{s_2,t} KL(p_{i, \tau}^{s_2}||p_{i,\tau}^t)\\
+ \tau^2 \beta_{s_2} \sum_i {\delta_{i, f}}^{s_2,s_2} KL(p_{i,\tau}^{s_1}||p_{i, \tau}^{s_2})\\
+ \tau^2 \beta_{s_2} \sum_i {\delta_{i, r}}^{s_2,s_1} KL(p_{i, \tau}^{s_2}||p_{i,\tau}^{s_1})\\
\end{multline*}
and the teacher loss is:
\begin{multline*}
L_{KD}^t = \alpha_t \sum_i L_{CE}(p_{i}^t, y_i)\\ + \tau^2 \beta_t \sum_i ( KL(p_{i,\tau}^t||p_{i, \tau}^{s_1}) + KL(p_{i,\tau}^t||p_{i, \tau}^{s_2})  )
\end{multline*}
\\
\subsection{Two Teachers One student (2T1S) loss functions } 
The teacher losses, in this case, are defined as follows:
\begin{multline*}
L_{KD}^{t_1} = \alpha_{t_1} \sum_i L_{CE}(p_{i}^{t_1}, y_i)  + \tau^2 \beta_{t_1} \sum_i KL(p_{i,\tau}^{t_1}||p_{i, \tau}^{s})
\end{multline*}
\begin{multline*}
L_{KD}^{t_2} = \alpha_{t_2} \sum_i L_{CE}(p_{i}^{t_2}, y_i)  + \tau^2 \beta_{t_2} \sum_i KL(p_{i,\tau}^{t_2}||p_{i, \tau}^{s})
\end{multline*}
and the student loss is:
\begin{multline*}
L_{KD}^{s} = \alpha_{s} \sum_i L_{CE}(p_{i}^{s}, y_i)\\
+ \tau^2 \beta_{s} \sum_i {\delta_{i, f}}^{s,t_1} KL(p_{i,\tau}^{t_1}||p_{i, \tau}^s)\\
+ \tau^2 \beta_{s} \sum_i {\delta_{i, r}}^{s,t_1} KL(p_{i, \tau}^{s}||p_{i,\tau}^{t_1})\\
+ \tau^2 \beta_{s} \sum_i {\delta_{i, f}}^{t_2,s} KL(p_{i,\tau}^{s}||p_{i, \tau}^{t_2})\\
+ \tau^2 \beta_{s} \sum_i {\delta_{i, r}}^{s,t_2} KL(p_{i, \tau}^{s}||p_{i,\tau}^{t_2})\\
\end{multline*}

\section{T-SNE Visualizations}
\begin{figure}[h]
    \begin{minipage}{0.23\textwidth}
    \includegraphics[width=\linewidth]{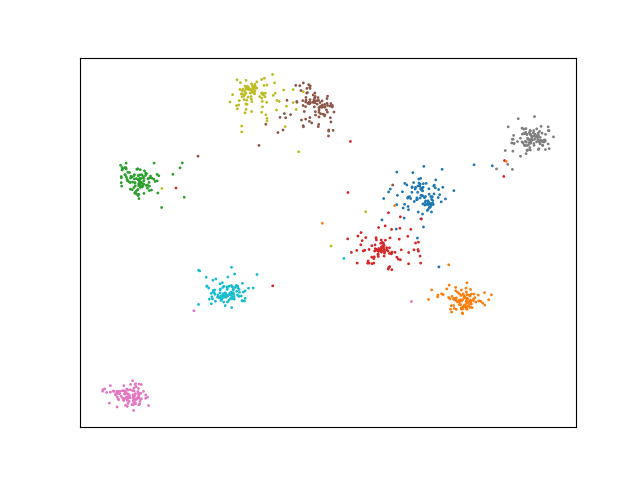}
    \centering{(a) ResNet32x4 Teacher with BD-KD}
    \end{minipage}
    \hfill
    \begin{minipage}{0.23\textwidth}
    \includegraphics[width=\linewidth]{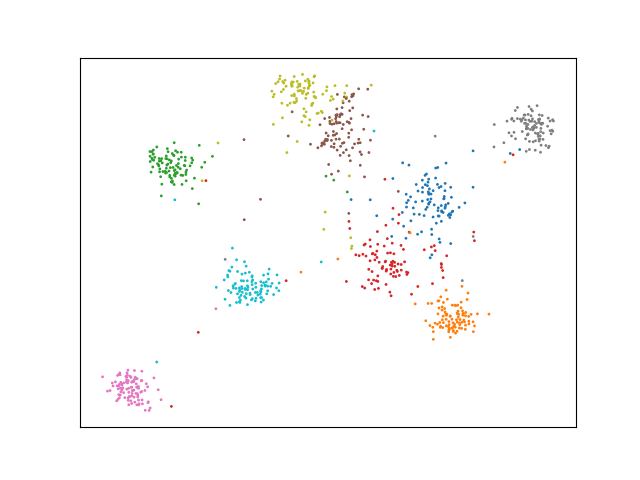}
    \centering{(b) ResNet8x4 Student with BD-KD}
    \end{minipage}
    \hfill
    \begin{minipage}{0.23\textwidth}
    \includegraphics[width=\linewidth]{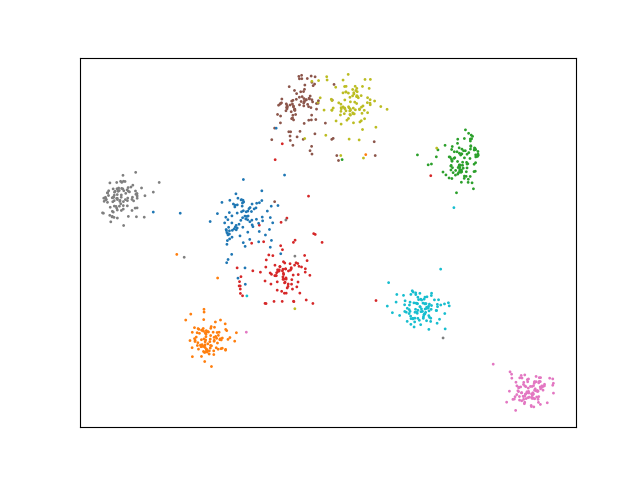}    
    \centering{(c) ResNet32x4 Teacher with SwitOKD}
    \end{minipage}
    \hfill
    \begin{minipage}{0.23\textwidth}
    \includegraphics[width=\linewidth]{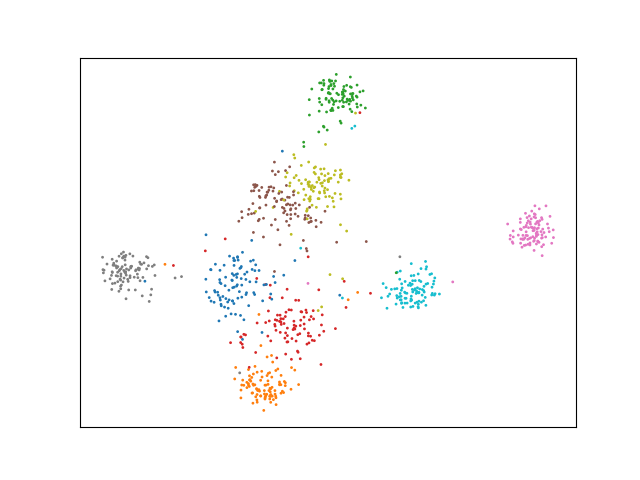}
    \centering{(d) ResNet8x4 Student with SwitOKD}
    \end{minipage}
    \caption{TSNE visualization of the penultimate feature layer of both teacher (ResNet32x4) and student (ResNet8x4) networks on CIFAR-100. We randomly picked and visualized 10 classes. }
    \label{fig:tsne1}
\end{figure}
\begin{figure}[h]
    \begin{minipage}{0.23\textwidth}
    \includegraphics[width=\linewidth]{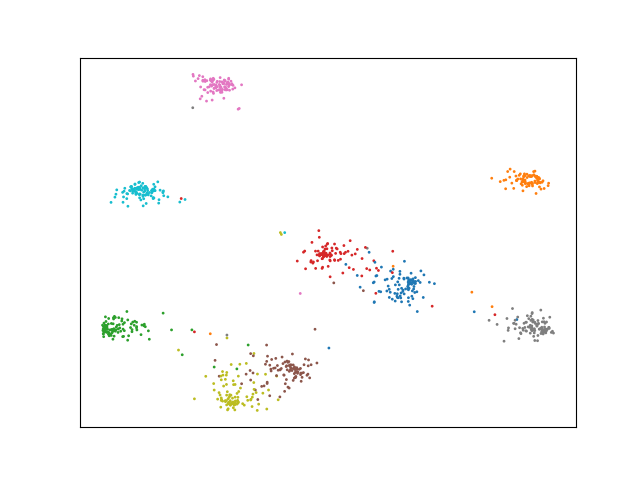}
    \centering{(a) VGG13 Teacher with BD-KD}
    \end{minipage}
    \hfill
    \begin{minipage}{0.23\textwidth}
    \includegraphics[width=\linewidth]{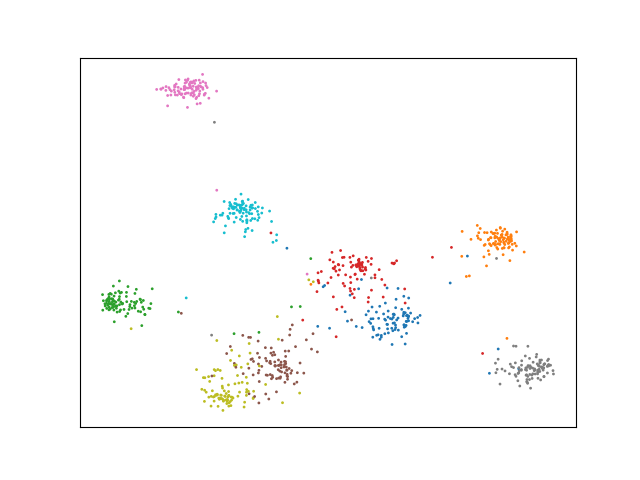}
    \centering{(b) VGG8 Student with BD-KD}
    \end{minipage}
    \hfill
    \begin{minipage}{0.23\textwidth}
    \includegraphics[width=\linewidth]{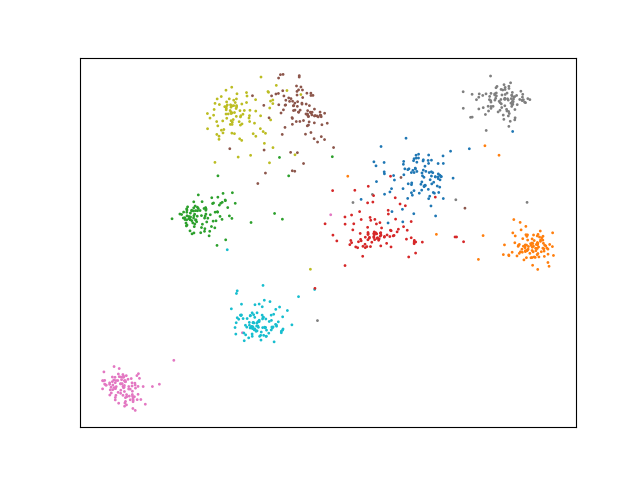}    
    \centering{(c) VGG13 Teacher with SwitOKD}
    \end{minipage}
    \hfill
    \begin{minipage}{0.23\textwidth}
    \includegraphics[width=\linewidth]{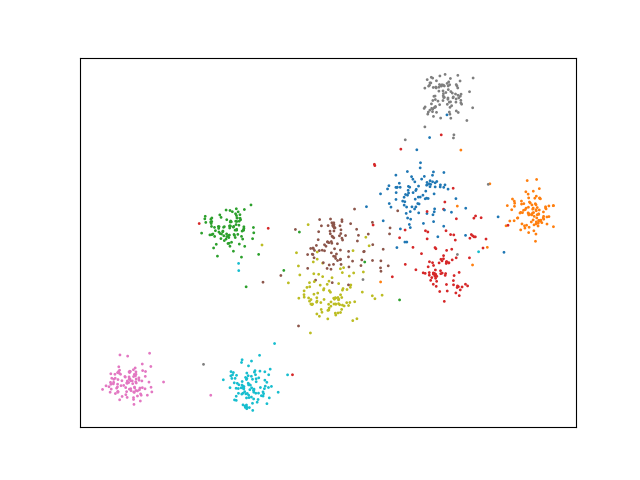}
    \centering{(d) VGG8 Student with SwitOKD}
    \end{minipage}
    \caption{TSNE visualization of the penultimate feature layer of both teacher (VGG13) and student (VGG8) networks on CIFAR-100. We randomly picked and visualized 10 classes.}
    \label{fig:tsne2}
\end{figure}

We provide in Figure \ref{fig:tsne1}  the TSNE visualization of the final layer (before the classifier) for the teacher network ResNet32x4 and student network ResNet8x4 trained on CIFAR-100. We also illustrate in Figure \ref{fig:tsne2} the TSNE visualization on CIFAR-100 with teacher network VGG13 and student network VGG8. We randomly picked 10 classes and visualized them.
We observe that with BD-KD the classes are more separable when compared to the same teachers trained with SwitOKD. Similarly, for the student network, we could see a better separation of classes using BD-KD (Figure \ref{fig:tsne2}) when compared to the same student architecture. 

\section{Frequently Asked Questions}
\subsection{Why use online distillation to improve calibration?}
Online distillation, where both teacher and student networks engage in mutual training, offers significant advantages over offline methods, as demonstrated by the SwitOKD paper. In online distillation, the student actively learns from the teacher and can continuously provide feedback signals to the teacher model. This interaction enables the teacher to adapt and update its guidance to the student based on recent information. In contrast, offline distillation typically involves a one-way transfer. 
Furthermore, this mutual real-time exchange of feedback between both models in KD is instrumental for calibration as shown in our experiments and ablation in the main manuscript. Our findings indicate that online KD techniques are better calibrated than offline KD. We also have shown that BD-KD is the best calibrated among online techniques, and simultaneously provides better performance accuracy.

\subsection{The predictive distribution of S matches that of T during training?}
The main objective of balancing the divergences is not to achieve a perfect match between the predictive distribution of the student network and that of the teacher network. Instead, it serves two key goals: Firstly, it aims to mitigate the accuracy gap during training, which is crucial in preventing disparities between the knowledge and confidence levels of the teacher and student models. To illustrate this, we included in our main submission visualizations of the behavior of the accuracy gap between teacher and student models. Additionally, balancing the divergences helps to align the uncertainty levels between the student and the teacher networks. This alignment is particularly important for maintaining a stable and effective knowledge transfer, even when an inherent capacity gap exists between two networks. These combined objectives enable the student to avoid aggressive mimicry of the teacher, driven by the substantial differences in confidence levels. 
A study \cite{muller2019does} has pointed out a noticeable gap between the predictive distribution of the student and the teacher in most distillation techniques. In light of this challenge, our proposed loss function is specifically designed to tackle this issue. Our loss function strives to bridge the gap making distillation more reliable.

\end{document}